# A CRITICAL ANALYSIS OF INTERNAL RELIABILITY FOR UNCERTAINTY QUANTIFICATION OF DENSE IMAGE MATCHING IN MULTI-VIEW STEREO

Debao Huang[1,2,3], Rongjun Qin[1,2,3,4*]

[1] Geospatial Data Analytics Laboratory, The Ohio State University, Columbus, USA – (huang.3918, qin.324)@osu.edu
[2] Department of Civil, Environmental and Geodetic Engineering, The Ohio State University, Columbus, USA
[3] Department of Electrical and Computer Engineering, The Ohio State University, Columbus, USA
[4] Translational Data Analytics Institute, The Ohio State University, Columbus, USA

**Commission IV, WG IV/2**

**ABSTRACT:**

Nowadays, photogrammetrically derived point clouds are widely used in many civilian applications due to their low cost and flexibility in acquisition. Typically, photogrammetric point clouds are assessed through reference data such as LiDAR point clouds. However, when reference data are not available, the assessment of photogrammetric point clouds may be challenging. Since these point clouds are algorithmically derived, their accuracies and precisions are highly varying with the camera networks, scene complexity, and dense image matching (DIM) algorithms, and there is no standard error metric to determine per-point errors. The theory of internal reliability of camera networks has been well studied through first-order error estimation of Bundle Adjustment (BA), which is used to understand the errors of 3D points assuming known measurement errors. However, the measurement errors of the DIM algorithms are intricate to an extent that every single point may have its error function determined by factors such as pixel intensity, texture entropy, and surface smoothness. Despite the complexity, there exist a few common metrics that may aid the process of estimating the posterior reliability of the derived points, especially in a multi-view stereo (MVS) setup when redundancies are present. In this paper, by using an aerial oblique photogrammetric block with LiDAR reference data, we analyze several internal matching metrics within a common MVS framework, including statistics in ray convergence, intersection angles, DIM energy, etc. We associate these metrics to the per-point errors evaluated through LiDAR reference data and discuss their potential contributions in estimating internal reliabilities of point clouds derived from DIM algorithms. The experimental results show that ray convergence and DIM energy are relevant indicators for the accuracy of the generated point clouds. Initial investigation shows that these two indicators could be further utilized to infer the measurement errors without reference data, which could potentially estimate the reliabilities of point clouds through error propagation.

## 1. INTRODUCTION

Multi-view stereo (MVS) has been a hot research topic in the field of photogrammetry and computer vision for decades. As a key component in 3D reconstruction, MVS takes a set of oriented images normally estimated by photogrammetry or Structure from Motion (SfM) algorithms as the input, and then establishes the dense correspondences across the images, which could be further triangulated to point clouds. MVS provides a convenient and cost-effective means of scene modeling as compared to other approaches such as LiDAR. Therefore, MVS is commonly used as a critical component for many applications, such as 3D reconstruction and mapping (Elhashash and Qin, 2022; Hu and Mordohai, 2012; Stathopoulou et al., 2023; Xu et al., 2022), autonomous navigation (Heng et al., 2019; Kendall, 2019), robotics (Song et al., 2021), and augmented reality (Zhang et al., 2019). Four types of MVS algorithms are given in (Seitz et al., 2006), including voxel-based methods, surface evolution-based methods, feature point growing-based methods, and depth-map fusion-based methods. Among these, the depth-map fusion-based methods have the advantages of finer geometry and scalability, which are used in our experiments. There are several internal metrics derived from depth-map fusion-based MVS algorithms, such as ray convergence, intersection angle, dense image matching (DIM) energy, etc. In this work, we analyze these metrics to understand how they affect the accuracy of the photogrammetric point clouds.

Generally, the quality of generated point clouds is evaluated by measuring the accuracy against reference data, e.g., LiDAR point clouds. However, such reference data may be unavailable at an evaluated region, and not knowing the fidelity of point clouds may disqualify photogrammetric point clouds for many application scenarios, such as point targeting and navigation. Therefore, estimating the reliability of the generated point clouds is critical to downstream applications in areas where reference data are not available. The internal reliability of camera networks has been well studied through Bundle Adjustment (BA) using the Gauss-Markov Theorem (Thompson et al., 1966), which plays an important role in determining the precision of the final triangulated 3D points. However, determining the measurement errors of DIM and propagating measurement errors to the 3D points are still open problems. There are several works focusing on the measurement errors. The works presented by Kuhn et al. (2017) estimated the disparity error by classifying disparities into different classes based on so-called total variation (TV) and learning the disparity error distribution from ground-truth data. Another work by Mundy and Theiss (2021) estimated the measurement errors by running the semi-global matching (SGM) algorithms twice and assigning a probability value based on the consistency of the 3D points generated by the forward and reverse order of the stereo pair. The method presented by Mehltretter and Heipke (2021) estimated the disparity uncertainty using the cost curve by learning features from the cost volume. However, there are two challenges from these existing

---

| Metric | Strengths | Weaknesses |
|---|---|---|
| Properties of matching cost curve (Egnal et al., 2004; Hirschmüller et al., 2002; Veld et al., 2018; Zhang and Shan, 2001) | 1. Rich information and characteristics could be extracted from the cost curves. 2. The ability to reason about the distinctness. | 1. Hardly deal with large images since the cost volume is built upon pyramid instead of every pixel. 2. Confidence-based methods have no actual units to express the value. 3. Only in the context of stereo matching. |
| Consistency of dense matching between forward and reverse order of stereo pair (Egnal et al., 2004; Hu and Mordohai, 2012; Mundy and Theiss, 2021) | 1. Take image appearance and viewing angle differences into consideration. 2. Good at identifying occlusion and discontinuities. | 1. Easily result in equal confidence for most matches. 2. Confidence-based methods have no actual units to express the value. 3. Only in the context of stereo matching. |
| Disparity oscillation within an image patch (Kuhn et al., 2017; Rodarmel et al., 2019) | 1. Have certain unit interpretation. 2. Pixelwise prediction while considering the neighborhood information. | 1. Learn from ground truth data, meaning the generalization is limited. 2. Require ground truth disparity maps for learning. 3. Only in the context of stereo matching. |
| Learning-based methods (Kendall, 2019; Mehltretter and Heipke, 2021) | 1. Have certain unit interpretation. 2. Learning-based feature extraction captures more characteristics and informative features. | 1. Huge memory consumption for large images. 2. Require ground truth data for learning. |
| Ours | 1. Have certain unit interpretation. 2. In the contexts of both stereo matching and MVS. 3. Learning and predicting in a self-supervised manner. | 1. Do not leverage the rich information from the cost curves. 2. Potential difficulties in calibrating the metrics between stereo matching and MVS. |

**Table 1**. Strengths and weaknesses of the existing metrics and ours.

approaches: first, most of the existing approaches learned the error prediction from samples, which might lack generalization. Second, these error metrics are mostly based on a single stereo pair, while typical photogrammetric point clouds are carried out through MVS, where the information from multiple views should be, but rarely utilized in error predictions. To this end, we aim to predict the measurement errors in a self-supervised manner by leveraging the knowledge of stereo matching as well as the MVS framework comprehensively. Such inference requires a preliminary study on the metrics involved in the stereo matching and MVS processes, which is the goal of this work.

In this paper, we performed an initial investigation on indicators from an MVS framework that could potentially contribute to error prediction of the generated 3D point clouds: we first derived a few indicators within an MVS framework from both stereo and MVS perspectives, including viewing angles (composition of stereo pair), DIM energy, intersection angles, ray convergence (number of rays), etc. Then we performed a critical analysis to evaluate their correlation with the accuracy of the generated 3D point clouds against reference data. Finally, as an initial study, we investigated their potential to estimate the reliability of point clouds derived directly from DIM algorithms when reference data are not available. This is a work-in-progress report on initial results.

The rest of this paper is organized as follows: **Section 2** introduces the criteria for selecting the metrics in the MVS framework. **Section 3** presents the experimental results and analyses. **Section 4** concludes this work with the outlook for future work.

## 2. METHODOLOGY

The photogrammetric point clouds are generated by two main steps. Firstly, an image orientation process through incremental orientation estimation and BA, to estimate the camera poses and a sparse reconstruction can be generated as a by-product. Secondly, within the MVS framework, a typical dense matching algorithm, such as SGM, can be applied to find the dense correspondences and generate the depth maps for each image in the dataset. Existing methods leverage various types of information from the matching process, including cost curve, matching consistency, disparity smoothness, and learning-based cues. While they achieve more or less promising results in certain contexts, they are also limited by different factors as shown in **Table 1**. In our work, we define several internal metrics which are derived within the MVS framework to evaluate their relationship with the accuracy of the reconstruction. For the MVS framework, we consider the number of images observing and triangulating a 3D point, which we call it number of rays, as one of the internal metrics statistically related to ray convergence. Another internal metric used in this work is the median intersection angles of the stereo pairs formed by each image and all its neighboring images, which presents statistics about the intersection angles of the stereo pairs. As for the stereo matching process, we define the viewing angles and DIM energy as the internal metrics, as they are relevant to the theoretical precision of the 3D point, as well as the matching confidence. Our proposed metric combo from both stereo and MVS perspectives has its uniqueness: firstly, it leverages the advantages of both contexts. Secondly, it does not depend on learning from the ground truth data, making it more generalized to vast datasets. Thirdly, it can be expressed in specific metric units, e.g., in pixels. We also analyze and model the distribution of reprojection errors with respect to these metrics. This will help understand their roles in reliability analysis and error propagation of 3D points generated from the photogrammetric process.

In this section, we briefly describe the MVS framework in our experiments in **Section 2.1**. The internal metrics from both MVS and stereo matching perspectives are listed and explained in **Section 2.2** and **Section 2.3**. The modeling of the distribution of reprojection errors is introduced in **Section 2.4**.

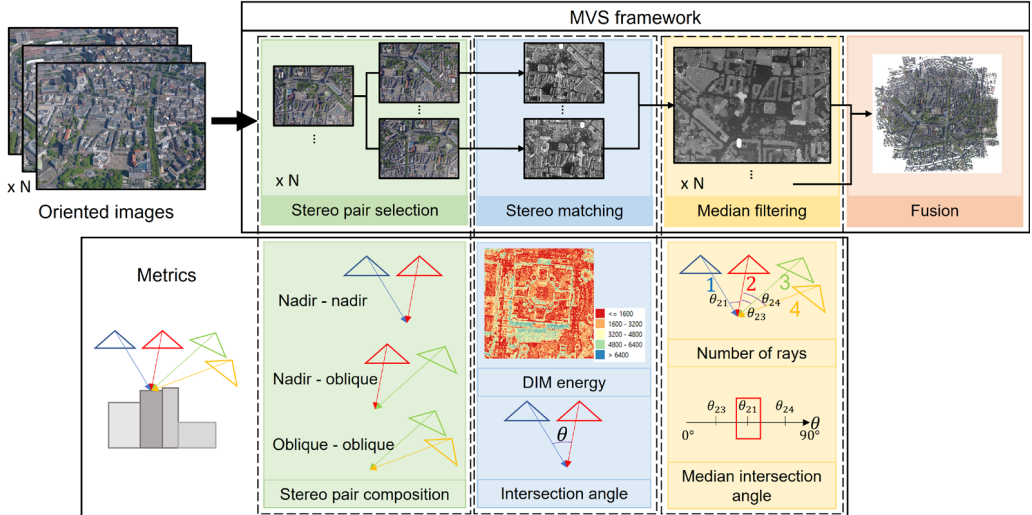

**Figure 1**. Overview of the MVS framework and metrics. Detailed description of this figure can be found in the text.

## 2.1 Overview of MVS Framework

An overview of MVS framework can be found in **Figure 1**, where stereo matching and MVS based filtering and fusion are two main steps of this process.

**Stereo Matching**. Given a stereo pair, both images are rectified and we use the Census-based SGM algorithm (Hirschmuller, 2007; Zabih[1] and Woodfill, 1994) to establish the dense correspondences, where the DIM energy is expressed as:

$$E = \sum_p \left( C(p, D_p) + \lambda \sum_{q \in N_p} S(D_p, D_q) \right), \quad (1)$$

where $C$ is the matching cost for pixel $p$ with a disparity value $D_p$. $S$ computes the smoothness of the neighborhood $N_p$ of pixel $p$. The relation between the cost term and smoothness term is adjusted by the coefficient $\lambda$. A hierarchical approach (Rothermel et al., 2012) is adopted to achieve memory efficiency. In the MVS framework, each row across the green and blue blocks in **Figure 1** can be regarded as an individual stereo matching process.

**MVS Based Filtering and Fusion**. A typical MVS framework is depicted in **Figure 1**. Given a set of oriented images, we first construct a set of stereo pairs for each image, as shown in the green block in **Figure 1**. For each input image $I_i$ in the dataset, we form a group of neighboring views $N(i)_j, j = 1, 2, \ldots, n$ based on certain criteria, e.g., camera poses and number of correspondences. In all the experiments we set $n$ to 10, thus each image finds at most 10 neighboring images to construct stereo pairs. For each stereo pair, we perform the stereo matching to find the dense correspondences and generate respective depth maps. As shown in the blue block in **Figure 1**, a set of $n$ depth maps $D(i)_j$ are generated for each image $I_i$ pairing with its $n$ neighboring views $N(i)_j$. For each pixel in the image $I_i$, it will be back projected to a 3D point only if its depth values exist in at least $k$ depth maps from $D(i)_j$, which we set to 2 in our experiments. As shown in the yellow block in **Figure 1**, the fused depth map $D_i$ for each image $I_i$ is generated by median filtering of the 3D points generated from all the observed depth maps (at least 2) from $D(i)_j$. The final photogrammetric point clouds are simply the merging results of 3D points derived from the fused depth maps of each image as shown in the pink block in **Figure 1**.

There are several metrics introduced for each block of the MVS framework. As shown in the bottom half of **Figure 1**, some of the metrics are involved in the stereo matching process (as outlined in green and blue blocks in **Figure 1**), while others are relevant to geometric configurations of the multi-view images (as outlined in the yellow block in **Figure 1**). The following subsections will discuss these metrics respectively.

## 2.2 Geometric Configurations of MVS as Metrics

Based on the extractable parameters from the MVS process, we consider analyzing the following metrics as potential indicators of the point uncertainty: 1) number of rays, which describes how many convergent rays a 3D point is triangulated from; 2) median intersection angle of these converging rays. Each metric is explained in the following subsections.

### 2.2.1 Number of Rays

In the MVS process, for each point cloud generated from the fused depth map of each image, a 3D point is retained only if at least two depth maps contain its depth value. In other words, at least three rays (each image itself and two neighboring views) are required. Since each image in the dataset is matched with at most 10 neighboring images, the number of rays is expected to range from 3 to 11 for a 3D point. Such a metric could be critical to the accuracy of the generated 3D point since robustness is more likely to be guaranteed for a high number of rays due to redundancy. Performing median filtering on points with a high number of rays is also expected to be closer to the ground truth position. In contrast, points with few rays could potentially suffer from unreliability due to unstable median filtering.

### 2.2.2 Median Intersection Angles

Since the 3D points are the fused results by median filtering of the stereo pairs, median intersection angles reflect the statics of intersection angles of all the stereo pairs. A median intersection angle that is too small or too large indicates the intersection angles of all the stereo pairs are possibly not ideal for triangulation. If the intersection angle is too small, the camera rays become more parallel, which makes triangulation more sensitive to noises as it is an ill-posed problem. Particularly, it affects the accuracy in the depth direction. On the other hand, the intersection angle being too large can lead to a huge appearance difference, which makes the dense matching even more challenging. In addition, it also introduces the occlusion problem. Thus, a good intersection angle is critical for triangulation and the median intersection angle is worth further analysis.

### 2.3 Stereo Matching Metrics

In the MVS framework, the SGM algorithm is performed for each stereo pair. Therefore, several internal metrics from the stereo matching process can be explored: 1) viewing angles of the images in the stereo pair, which can construct different types of stereo pairs; 2) DIM energy, which describes the dense matching quality. Each metric is explained in the following subsections.

#### 2.3.1 Viewing Angles

Considering the aerial dataset consists of both nadir and oblique images, three types of stereo pairs can be possibly constructed: two nadir images, two oblique images, one nadir image with one oblique image. Stereo pairs constructed by two images with either similar viewing angles or not might have great impacts on the accuracy of dense matching as well as the intersection angles. Thus, it is interesting to analyze the effect of the composition of stereo pairs on the accuracy of generated 3D points. Additionally, the distance between the points generated from single stereo pair and the point clouds from the median filtering can be measured to analyze what types of stereo pairs tend to generate more deviated points that not only get filtered out by median filtering but also affect the results of median filtering.

#### 2.3.2 DIM Energy

This metric reflects the quality of dense matching directly. Generally, low and smooth DIM energy around the pixels reflects high confidence of successful matches. On the contrary, dense matching on challenging areas is expected to have high DIM energy due to the displacement of correspondences and thus leads to poor triangulation.

### 2.4 MVS Reprojection Errors

For a pixel $p$ in the image $I_i$, if the ground truth of the associated 3D point $X_p$ is known, the reprojection error of the neighboring view $I_j$ can be computed as:

$$R(I_i, I_j)_p = \|P(X_p, I_j) - \lambda(p, I_j)\|_2 , \qquad (2)$$

where $P(X_p, I_j)$ is the projection of $X_p$ with respect to the neighboring image $I_j$, $\lambda(p, I_j)$ is the location of dense correspondence of $p$ in $I_j$ derived by SGM algorithms. If the reprojection error of $I_i$ itself is small (i.e., smaller than 1 pixel), we can approximate measurement errors using reprojection errors of $I_j$ since measurement errors indicate the displacement of the dense correspondences in the neighboring images. However, if no ground truth data are available, there is no way to compute the reprojection errors. In this case, the reprojection errors can be approached using 3D points that are most likely to be accurate based on MVS metrics that are highly correlated to accuracy. By modeling the distribution of reprojection errors associated with highly correlated stereo matching metrics using these "accurate" 3D points, the reprojection errors can be inferred for other points based on the same metric, thus their measurement errors can be also estimated. To choose an appropriate distribution model for the reprojection errors, we first look at the components of it. The reprojection error $R$ contains residuals in both $X$ and $Y$ directions, which we assume normal distributions with different means and standard deviations for each component. Thus, $R$, which is computed as the squared root of the residual sum of squares in $X$ and $Y$ directions, its distribution model can be approximated by a general Gamma distribution.

## 3. EXPERIMENTAL RESULTS

Experiments were performed to evaluate the potential of internal metrics described in **Section 2** for indicating the accuracy of point clouds against LiDAR reference data. The contribution of the potential indicators to the estimation of the reliability of point clouds derived from the DIM algorithm was further analyzed. In all experiments, the 3D point clouds were generated through the MVS framework described in **Section 2**. 3D points triangulated directly from each stereo pair were used to evaluate the metrics associated with the stereo matching process, while 3D points generated by median filtering with at least three rays were used to evaluate the metrics corresponding to the MVS process. Prior to the evaluation, the generated point clouds were registered to the LiDAR point clouds using ICP algorithm (Besl and McKay, 1992) implemented in the open-source software CloudCompare (Girardeau-Montaut, 2022). The accuracy against LiDAR point clouds was then measured by computing the mean absolute distance (or mean absolute error, MAE) and each point has its per-point error.

**Section 3.1** introduces the imagery dataset used in the experiments and the LiDAR reference data. The experimental results and analyses are described in detail in **Section 3.2**.

### 3.1 Dataset

Dortmund airborne imagery dataset and aerial LiDAR point clouds were used in our experiments. Dortmund dataset (Nex et al., 2015) contains nadir and oblique airborne images acquired by the airborne oblique image system (IGI PentaCam) with five camera heads. LiDAR point clouds collected by Aerial Laser Scanner (ALS) were used for evaluation of accuracy. **Table 2** shows the specification of the datasets. **Figure 2** shows the photogrammetric point clouds generated by the MVS framework and the LiDAR point clouds covering the same area.

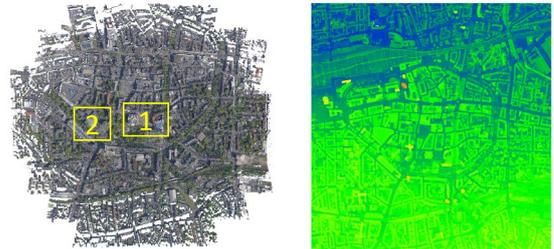

**Figure 2**. Left: overview of the photogrammetric point clouds generated from Dortmund imagery dataset. Yellow boxes indicate two selected sub-regions. Right: overview of LiDAR point clouds.

| Imagery dataset | |
|---|---|
| Number of images | 16 (N), 43 (O) |
| Camera system | PentaCam IGI |
| Image size | 6132 × 8176 pixels (N) <br> 8176 × 6132 pixels (O) |
| GSD | 10 cm (N), 8 – 12 cm (O) |
| Overlap | 75%/80% (N), 80%/80% (O) |
| LiDAR dataset | |
| Density | 10 pts/m$^2$ |

**Table 2**. Specifications of the datasets. "N" refers to nadir images, "O" refers to oblique images. "GSD" refers to ground sampling distance. The overlap contains along/across-track directions.

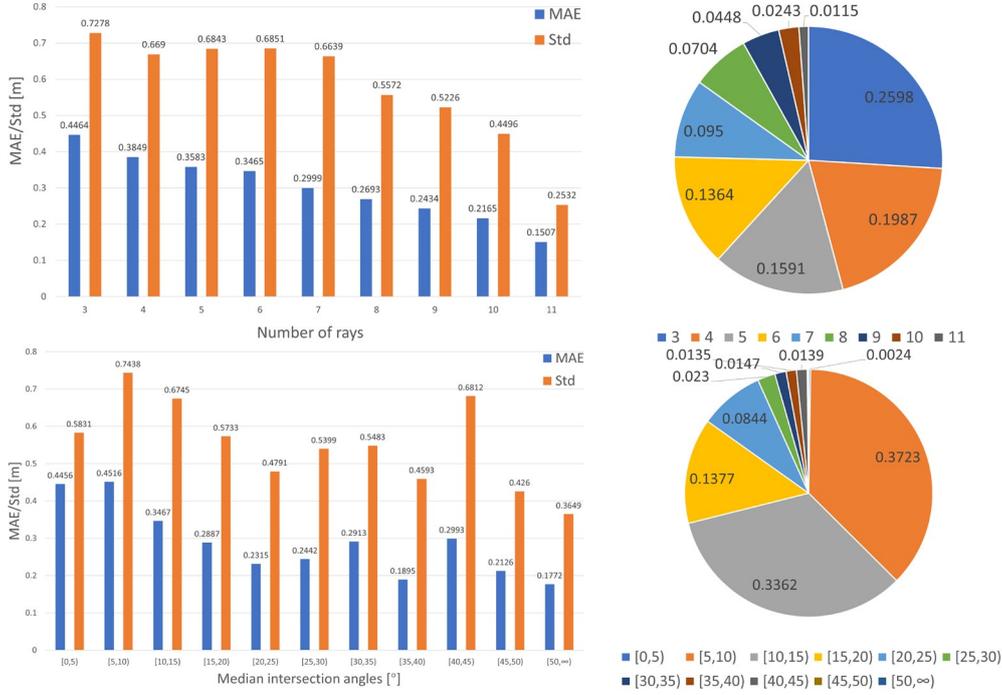

**Figure 3**. Left column: comparison of the effects of different numbers of rays and median intersection angles on the accuracy of generated point clouds. Right column: proportions of different number of rays and median intersection angles. For the pie figure of median intersection angles, the proportion values of two thinnest slices [40, 45) and [50, ∞) are not shown for better visualization purpose. The order of slices in the pie figures is sorted clockwise by the legends.

### 3.2 Experimental Results

#### 3.2.1 Evaluation of MVS Indicators

The first experiment was performed on point clouds generated by the MVS process as discussed in **Section 2**. We registered the point clouds to the LiDAR point clouds and measured the accuracy. For each 3D point, a per-point error was then associated with matching metrics in the MVS framework to analyze their correlation. The point clouds were split based on different value ranges of each matching metric, so MAE as well as standard deviation can be obtained respectively. The goal is to find the potential indicators for the accuracy of point clouds.

**Figure 3** shows the statistical results of accuracy with respect to different numbers of rays and median intersection angles. Firstly, a decreasing trend of MAE and standard deviation can be observed as the number of rays increases, indicating points with high accuracy tend to have multiple rays. The number of such points (e.g., with more than 5 rays) is more than 30% of all the points, which is sufficient for the number of rays to serve as a potential indicator for accuracy. As for median intersection angles, except for a few outliers with extremely small or large angles (e.g., smaller than 5 degrees or larger than 45 degrees), we found that MAE and standard deviation gradually decrease as median intersection angles become larger but start to fluctuate once the angle is larger than 25 degrees. 3D points with median intersection angles ranging from 20 to 25 degrees achieve the best accuracy. The number of points with median intersection angles falling in this range is also sufficient (13.77%) for median intersection angles to serve as another potential indicator of accuracy.

The above analysis presents two potential indicators for the accuracy of generated point clouds. We further investigated their distinctness by separating points into groups with high and low accuracy. Two sub-regions were extracted from the entire point clouds for evaluation, as shown in **Figure 2**. The extracted point clouds were split by per-point errors with a threshold of 0.5 m. For the points with high and low accuracies, the distributions of the number of rays and median intersection angles are shown in **Figure 4**. Two peaks can be clearly distinguished for the distributions of the number of rays for points with large/small errors in both sub-regions. On the contrary, the peaks of distributions of median intersection angles cannot be well separated for points with large/small MAE in both sub-regions. The results indicate that number of rays can be served as a better indicator to the accuracy, while median intersection angles lack such distinctness. Visual results in **Figure 5** reflect this finding: the areas with high errors are mostly matched with the areas with a low number of rays, while there is no strong pattern between the error distribution and the distribution of median intersection angles.

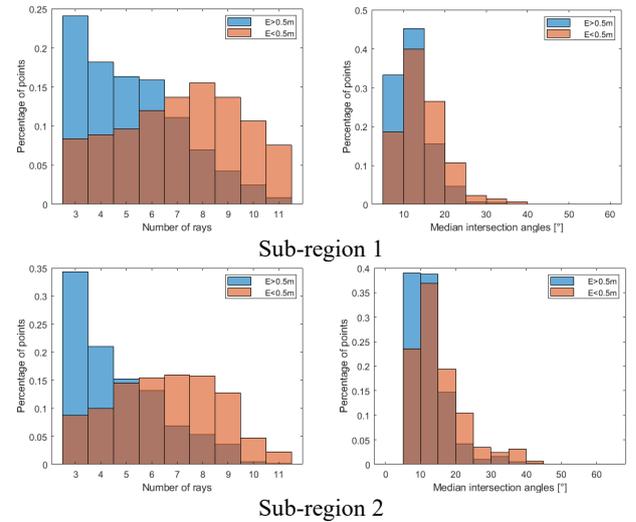

**Figure 4**. Histogram of number of rays and median intersection angles in areas with small errors (< 0.5 m) and areas with large errors (> 0.5 m) for both sub-regions.

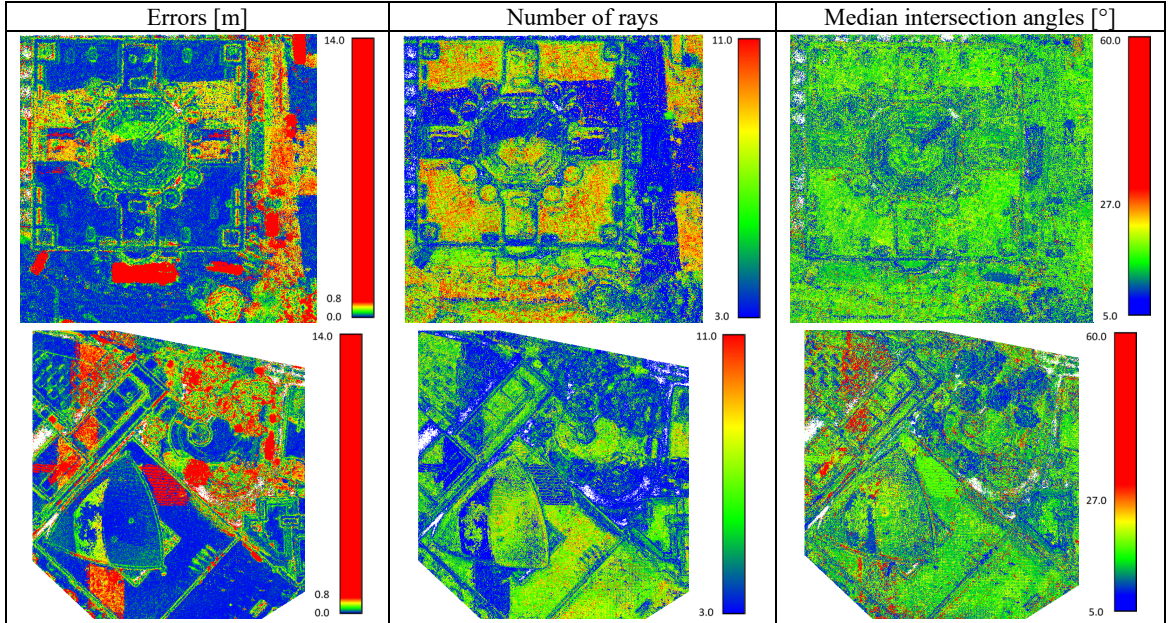

**Figure 5**. Distribution of per-point errors, number of rays, and median intersection angles of the generated point clouds in region 1 (first row) and region 2 (second row). The scale of color rendering is adjusted for better visualization purpose.

### 3.2.2 Evaluation of Stereo Matching Indicators

We further evaluated the potential indicators in the stereo matching process by generating and comparing 3D points directly from each stereo pair to the LiDAR point clouds. The 3D points generated by stereo pairs were also compared with the 3D points generated by the MVS process to analyze what type of stereo pairs tends to generate more outliers filtered out by median filtering. Registration as a pre-processing step used the same transform matrix as in the evaluation of MVS indicators.

Considering that the dataset contains both nadir and oblique images, the composition of the stereo pair was first evaluated with all possible combinations (i.e., nadir/nadir, nadir/oblique, oblique/oblique). **Table 3** shows the statistical results compared to the 3D point clouds generated by MVS. Stereo pair which consists of both nadir images generates 3D points closest to 3D points generated by MVS and the standard deviation is the smallest. The points generated from the stereo pair of one nadir image and one oblique image have the largest distances and twice the standard deviation, which are more likely to be outliers during the median filtering process in MVS since they are most deviated. The distribution of intersection angles as shown in **Figure 6** explains the reason: only the pairs with one nadir image and one oblique image have intersection angles larger than 35 degrees. Large intersection angle introduces more matching difficulties including appearance difference and occlusion, thus the 3D points generated by such pairs are more possible to become outliers.

The point clouds generated by the stereo pairs were also compared to the LiDAR point clouds. As shown in **Table 3**, 3D points generated from stereo pairs of both nadir images or one nadir image and one oblique image have similar MAE, while the 3D points generated from stereo pairs consisting of both oblique images have the worse MAE. Such results can be explained from the distribution of intersection angles of different stereo pairs as shown in **Figure 6**: the stereo pairs with both oblique images tend to have more small intersection angles, and almost fail to have intersection angle that falls in the range of 20 to 25 degrees, which is indicated as the best angle range in the first experiment. While stereo pairs consisting of one nadir image and one oblique image have all the points with large intersection angles (larger than 35 degrees), they also produce sufficient points with appropriate intersection angles. Therefore, using the composition of the stereo pair as an indicator might not be sufficient since the best combination (nadir/nadir) only contributes a small portion of all the possible stereo pairs and it is easily affected by the datasets.

| Stereo pair | MAE to 3D points by MVS | | MAE to LiDAR point clouds | |
|---|---|---|---|---|
| | Mean [m] | Std [m] | Mean [m] | Std [m] |
| Nadir/nadir | 0.2035 | 0.9100 | 0.2968 | 0.4327 |
| Nadir/oblique | 0.3484 | 1.8446 | 0.3054 | 0.6640 |
| Oblique/oblique | 0.2835 | 1.0485 | 0.4562 | 0.8205 |

**Table 3**. Distance between 3D points generated by stereo pairs with different combination and 3D points generated by MVS and LiDAR point clouds.

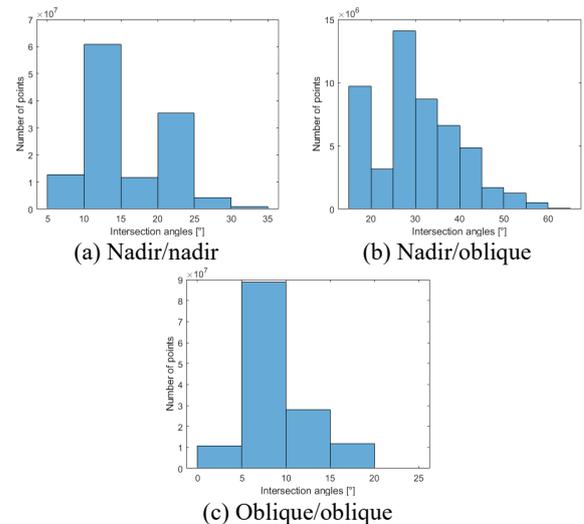

**Figure 6**. Distribution of intersection angles for different composition of stereo pairs.

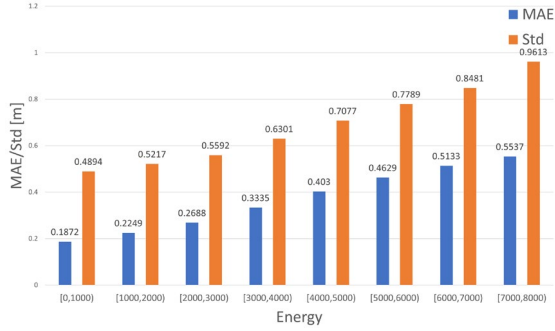

**Figure 7**. Comparison of the effect of different energy ranges on MAE and Std.

The relation between accuracy and DIM energy was evaluated for all the stereo pairs. **Figure 7** shows the comparison of the effect of different DIM energy ranges on the MAE and standard deviation. The results indicate that MAE is positively correlated to the DIM energy with the coefficient of determination $R^2$ 0.9938: the higher the DIM energy, the higher the MAE and standard deviation. Such a high correlation associates DIM energy with accuracy tightly, which makes it more reliable to estimate the reprojection errors.

The results of evaluation indicate that although the composition of stereo pair reflects the accuracy of point clouds to some extent, it has the similar problem as the metric of median intersection angles since it is subject to different datasets. On the other hand, DIM energy can be regarded as a good indicator for its high correlation with the accuracy of generated point clouds from stereo pairs.

### 3.2.3 Factors for Estimating Internal Reliability from DIM Algorithms

The above experiments were all conducted using LiDAR point clouds as reference data to evaluate the accuracy. In the case that such reference data are not available, inferring the reliability of point clouds from DIM algorithms can be critical. There are two key findings on the evaluation results of the indicators: 1) a large number of rays is critical for accurate 3D points; 2) DIM energy has a high correlation with the accuracy of the points. Based on the findings, number of rays and DIM energy could be leveraged to infer the reliability of point clouds by estimating the measurement errors and propagating to 3D points. This problem can be approached by using the multi-ray points as "ground truth" and estimating the reprojection errors based on the DIM energy, which is then used to approximate the measurement errors for all the points. The reliability of point clouds could be finally obtained through error propagation of the measurement errors. The reason is that a 3D point with a high number of rays is more likely to be accurate as "ground truth" as indicated in the above experiments. DIM energy can be used to estimate reprojection errors due to its high correlation with accuracy. Thus, an initial investigation was conducted by exploring the relationship between the DIM energy and the reprojection errors between the projection of the multi-ray points and the correspondences derived from the DIM algorithms.

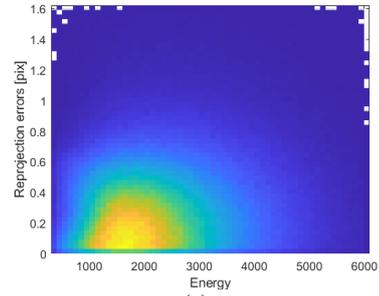
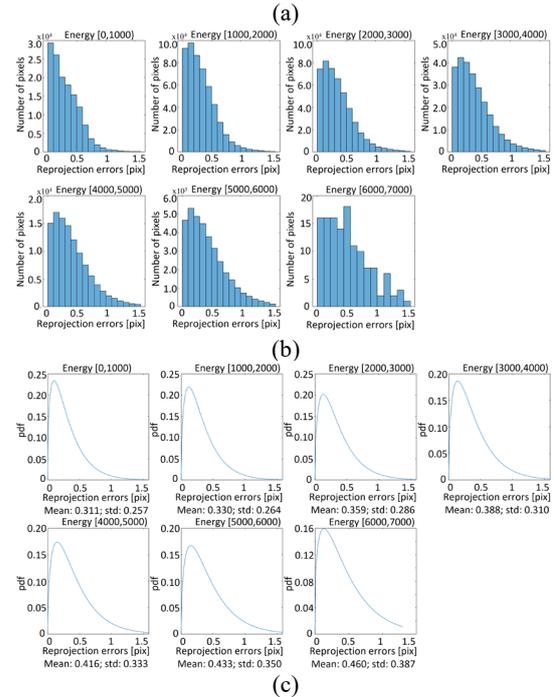

**Figure 8**. (a) Joint distribution of DIM energy and reprojection errors. (b) Histogram of reprojection errors under different DIM energy ranges. (c) Curve fitting of the histogram using Gamma distribution model.

**Figure 8(a)** shows the joint distribution of DIM energy and reprojection errors for one stereo pair. Since the selected points have a high number of rays, they tend to have low DIM energy and small reprojection errors (located in the bright yellow region in **Figure 8(a)**). The DIM energy was further categorized into different bins with the size of 1000 to compute the distribution of reprojection errors within different ranges of DIM energy, as shown in **Figure 8(b)**. It can be seen that as the DIM energy increases, the distribution of reprojection errors becomes dispersed. Based on the formulation of reprojection errors, we used the Gamma distribution model to approximate the distribution of reprojection errors, as depicted in **Figure 8(c)**. The statistical results show that the mean and standard deviation of reprojection errors increase as the DIM energy increases, which accordingly reflects the histograms in **Figure 8(b)**. Once the relationship between the reprojection errors and DIM energy is obtained using the multi-ray points, we can estimate the reprojection errors for other points based on their DIM energy. The estimation of measurement errors and error propagation can be further explored in future work.

## 4. CONCLUSIONS

In this paper, we performed a series of experiments to analyze the potential of several matching metrics within the MVS framework as indicators for the accuracy of the generated point clouds. An aerial imagery dataset consisting of nadir and oblique images was used to generate the photogrammetric point clouds, evaluated with aerial LiDAR point clouds as reference data. The results of experiments show that from the perspective of MVS process, number of rays is a good and distinct indicator of accuracy. Particularly, 3D points with more than 5 rays have a high possibility of being accurate. As for the median intersection angles, we found that for a typical aerial imagery dataset, 20 – 25 degrees are necessarily good angles to generate 3D points with good accuracy. However, it does not have strong distinctness as the number of rays in terms of distinguishing accurate and inaccurate points and it may vary with different datasets. From the perspective of stereo matching process, the DIM energy can be served as another indicator due to its high correlation with accuracy. We also found that stereo pairs consisting of both nadir images tend to produce points with better accuracy than those consisting of at least one oblique image. Additionally, as an initial investigation, we explored the potential of using the qualified indicators to infer the point reliability directly from the DIM algorithm when reference data are not available. Based on the findings on the indicators, 3D points with multiple rays (i.e., more than 5 rays) can be regarded as "ground truth" as they are more likely to be accurate. Using these points, we were able to model the reprojection errors with respect to the DIM energy by a Gamma distribution, which can further infer the measurement errors, thus the final 3D point reliability through error propagation. Further study can be focused on the quantification of such reliability and error propagation, and the quantitative comparison with other existing methods which rely on learning from ground truth samples.

## ACKNOWLEDGEMENT

This work was partially supported by the Office of Naval Research (ONR, Award No. N00014-20-1-2141 & N00014-23-1-2670). The authors would like to acknowledge the provision of the Dortmund dataset by ISPRS and EuroSDR.